\documentclass{article}
\usepackage{graphicx}

\begin{document}
\title{Wavelet and Curvelet Moments for Image Classification: Application 
to Aggregate Mixture Grading} 

\author{Fionn Murtagh \\
Department of Computer Science, \\
Royal Holloway, University of London, \\
Egham, Surrey TW20 0EX, England \\ 
fmurtagh@acm.org \\
 and \\
Jean-Luc Starck \\
Service d'Astrophysique \\
Centre d'Etudes de Saclay \\
Ormes des M\'erisiers \\
91191 Gif-sur-Yvette Cedex, France \\
jstarck@cea.fr}
\maketitle

\begin{abstract}
We show the potential for classifying images of mixtures of aggregate, 
based themselves on varying, albeit well-defined, sizes and shapes, in order 
to provide a far more effective approach compared to the classification of
individual sizes and shapes.  
While a dominant (additive, stationary) Gaussian noise component in image 
data will ensure that wavelet coefficients are of Gaussian distribution,
long tailed distributions (symptomatic, for 
example, of extreme values)  may well hold  in practice for 
wavelet coefficients.  Energy (2nd order moment) has often been used
for image characterization for image content-based retrieval, 
and higher order moments may 
be important also, not least for capturing long tailed distributional
behavior.  In this work, we assess 2nd, 3rd and 4th order moments of 
multiresolution transform -- wavelet
and curvelet transform -- coefficients as features.
As analysis methodology, taking account of image types, multiresolution 
transforms, and moments of coefficients in the scales or bands, we use
correspondence analysis as well as k-nearest neighbors supervised 
classification.  
\end{abstract}


\section{Image Grading as a Content-Based Image Retrieval Problem}

Physical sieves are used to classify crushed stone based on size and 
granularity.  Then mixes of aggregate are used.  We directly address 
the problem of classifying the mixtures, and we assess the 
algorithmic potential of this approach which has considerable industrial
importance.

The success of content-based image finding and retrieval is most
marked when the user's requirements are very specific.  An example of 
a specific application domain is the grading of engineering materials. 
Civil engineering construction {\em aggregate} sizing is carried 
out in the industrial context by passing the 
material over sieves or screens of particular sizes. 
Aggregate is a 3-dimensional material (images are shown later in this
article) and as such need not necessarily meet the screen 
aperture size in all directions so as to pass through that screen. The 
British Standard and other specifications 
suggest that any single size aggregate may contain a 
percentage of larger and smaller sizes, the magnitude of this percentage 
depending on the use to which the aggregate is to be put.
An ability to measure the size and shape characteristics of an aggregate 
or mix of aggregate, ideally quickly, is desirable to 
enable the most efficient use to be made of the aggregate and binder 
available.
This area of application is an ideal one for image content-based matching
and retrieval, in support of automated grading.  Compliance with 
mixture specification is tested by means of match against an image database
of standard images, leading to an automated ``virtual sieve''.  
Previous work includes Murtagh et al.\ (2005a, 2005b).

In this work we do not seek to discriminate as such between particles
of varying sizes and granularities, but rather to directly classify 
mixtures.  Our work shows the extent to which we can successfully address 
this more practical and operational problem.  As a ``virtual sieve'' 
this classification of mixtures is far more powerful than physical 
sieving which can only handle individual components in the mixtures.  

In section \ref{sect2} we will review the properties of multiresolution 
transform coefficients, given our planned use of these coefficients to 
discriminate between images and thus support image retrieval and/or grading.
In section \ref{sect3} we carry out a detailed study to assess the 
moments of multiscale transforms, wavelet and curvelet transforms, and 
use of images of different smoothness and edginess characteristics.
In section \ref{sect4} we carry this work further into a practical 
domain of application of image grading, viz.\ that of the civil engineering
construction materials.

\section{Distributions of Multiresolution Coefficients}
\label{sect2}

\subsection{Gaussian Distribution of Wavelet Coefficients}

Noise filtering from a wavelet transformed image, 
based on a Gaussian model, is highly developed in theory and practice.  
The monographs 
Starck et al.\ (1998a) and Starck and Murtagh (2006) study in great detail
wavelet-based noise filtering.  
Unlike e.g.\ smoothness criteria in noise filtering (Donoho and Johnstone, 
1995), our perspective in this work has been towards, or based on,
definable noise models in the data.  
Such noise filtering is generalized there 
for the Poisson case, through variance stabilization.  Alternative 
noise thresholding approaches are studied for other distributions, 
including  small count Poisson and Rayleigh.  The noise filtering 
is also studied in the perspective of optimal image restoration (see
Starck et al., 1998b), and is incorporated into image deconvolution.  

The Gaussian statistical model for images is particularly appropriate for 
the case of CCD (charge coupled device) image detectors, where Gaussian 
read-out noise is dominant.  

The Gaussian model leads to a close relationship between multiscale
(Shannon) entropy, wavelet energies, and variances (Starck et al., 1998b).
This lends weight to the use of the second order moment as an important 
multiscale feature.

The conclusion here is the following: when data is Gaussian distributed, 
then the modeling is very well understood.  

\subsection{Variance or Energy of Multiresolution Coefficients}

The analysis of texture has used Markov modeling of spatial context (Cross 
and Jain, 1983) and a wavelet transform provides another way to 
incorporate local spatial relationships.  Unser (1995) used co-occurrence 
matrices and concluded that second order statistics may be best for 
segmentation of microtextures.  The use of a wavelet transform 
for this purpose was first proposed by Mallat (1989). Scheunders et al.\
(1998) discuss multiband 
features, which they use 
with 3-band color data.  In Livens et al.\ (1996), energy 
in different bands is used.  This does not provide image rotational 
invariance and for this a sum of energies over bands at a given resolution
scale is proposed.  
Fatemi-Ghomi (1997) uses window size related to resolution scale 
within which to define features, and she discusses adaptive window sizes.

\subsection{Long Tailed Distribution of Wavelet Coefficients}

In the general use of multiresolution transforms, 
it is well known that wavelet coefficients can be of long tailed distribution
(Belge et al., 2000; Buccigrossi and Simoncelli, 1999; Murtagh and 
Starck, 2003). 
Long tailed distributed data include data characterized by long range
interactions, long memory processes, fractal or multifractal or 
self-similar processes, multiplicative noise regimes (Anteneodo and 
Tsallis, 2003), and signals with burstiness, abrupt changes, and spikes
(Bezerianos et al., 2003).  Applications to thresholding are in Murtagh 
and Starck (2003) and Wang and Chung (2005).  
The L\'evy distribution characterizes many 
such cases: $L(x) \propto \left( 1 - (1-q) \frac{x}{\lambda} 
\right)^{1/(1-q)}$.
When parameter $q \rightarrow 1$, this {\em power} law approaches an 
{\em exponential} law:
$\exp \left( - \frac{x}{\lambda} \right)$, which 
typifies Boltzmann-Gibbs thermodynamics and Gaussian statistics.  

\subsection{Multiresolution Tsallis Entropy}

Starck et al.\ (1998b) and Starck and Murtagh (2006) used 
multiscale Shannon entropy for image filtering, showing how it 
is clearly related to the second order moment in the Gaussian case.  

In 1928 Hartley developed the entropy of equiprobable events, and 
in 1948 this was generalized by Shannon and widely applied as a basis
for the theory of communications.  As opposed to the coding objectives
based on events, a statistical mechanics objective based on system states was 
developed, such that the same Boltzmann-Gibbs-Shannon entropy results.
In 1960 R\'enyi generalized the recursive rather than linear estimation.
From 1988 onwards Tsallis developed a generalized form of entropy, which 
happens to differ from Shannon and R\'enyi entropies in being non-logarithmic, 
to cater for fractal and self-similar systems, i.e.\ systems where 
invariance across resolution scales is of importance.  (See Kaniadakis and 
Lissia, 2004.) The roots as such of Tsallis entropy  go back to 1970
(Abe and Rajagopal, 2000).
The Shannon or Boltzmann-Gibbs-Shannon entropy is: 
$ S = - \sum_i p_i \ln p_i $.
In the thermodynamics perspective, as opposed to the
event space view, $p_i$ is the probability that the system is found in 
the $i$th configuration.  

The non-extensive or 
Tsallis entropy, with parameter $q$, a positive real, is given by 
$
S^T_q = - \frac{1}{q-1} \left( \sum_i (p_i^q - p_i) \right) 
= - \frac{1}{q-1} \left( \sum_i p_i^q - 1 \right)
$
As for the Shannon entropy we may consider a constant of proportionality 
here, the Boltzmann constant, which we have set to 1. 

The Tsallis entropy has been proposed for long tailed data.  
Tsallis entropy of wavelet transformed data was used in Rosso et al.\ (2002).
An image thresholding approach was related to Tsallis entropy in Portes
de Albuquerque et al.\ (2004).   
Sporring and Weickert (1999) considered R\'enyi entropy in 
scale spaces: Tsallis entropy, as we have suggested above, may be 
more appropriate.  Tsallis
entropy was used on scale space transformed data by Tanaka et al.\ (1999).
In Costa et al.\ (2002) Tsallis entropy is applied to one-dimensional 
signals on a regular discrete range of resolution scales, and plotted, in
order to characterize biomedical data.  

\subsection{Distribution of Wavelet Coefficients in Practice: a Case Study}

In this section we will take an empirical standpoint, using a batch of 
images.  We ask: How do we assess Gaussianity or long tailedness of wavelet 
coefficients?

\begin{figure}[htbp] 
   \centering
   \includegraphics[width=6cm]{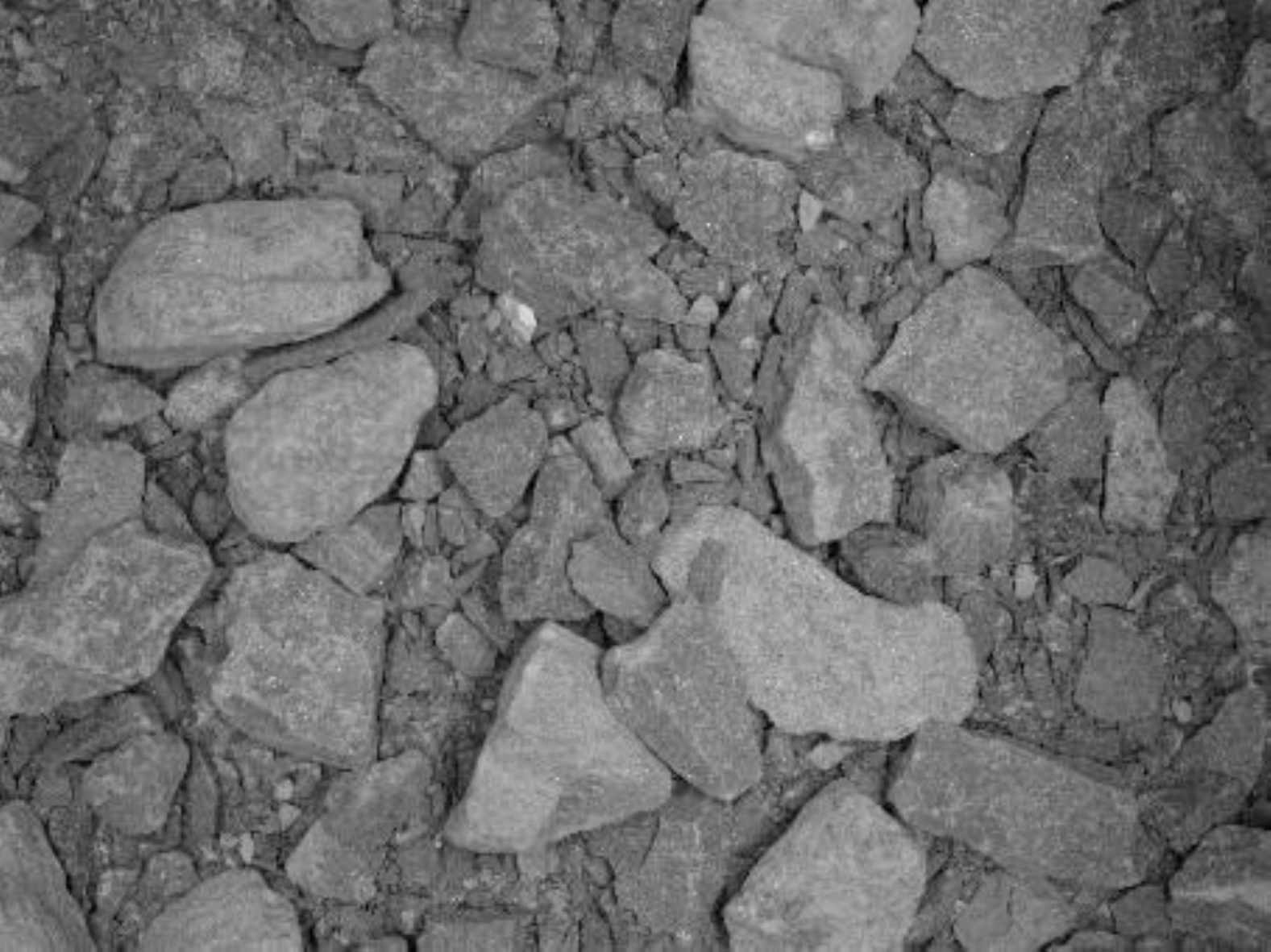} 
   \caption{An image of construction aggregate.  The properties of the 
aggregate are defined by size for larger pieces, and by granularity
for finer pieces.}
   \label{fig1}
\end{figure}

\begin{figure}[htbp] 
   \centering
   \includegraphics[width=7cm]{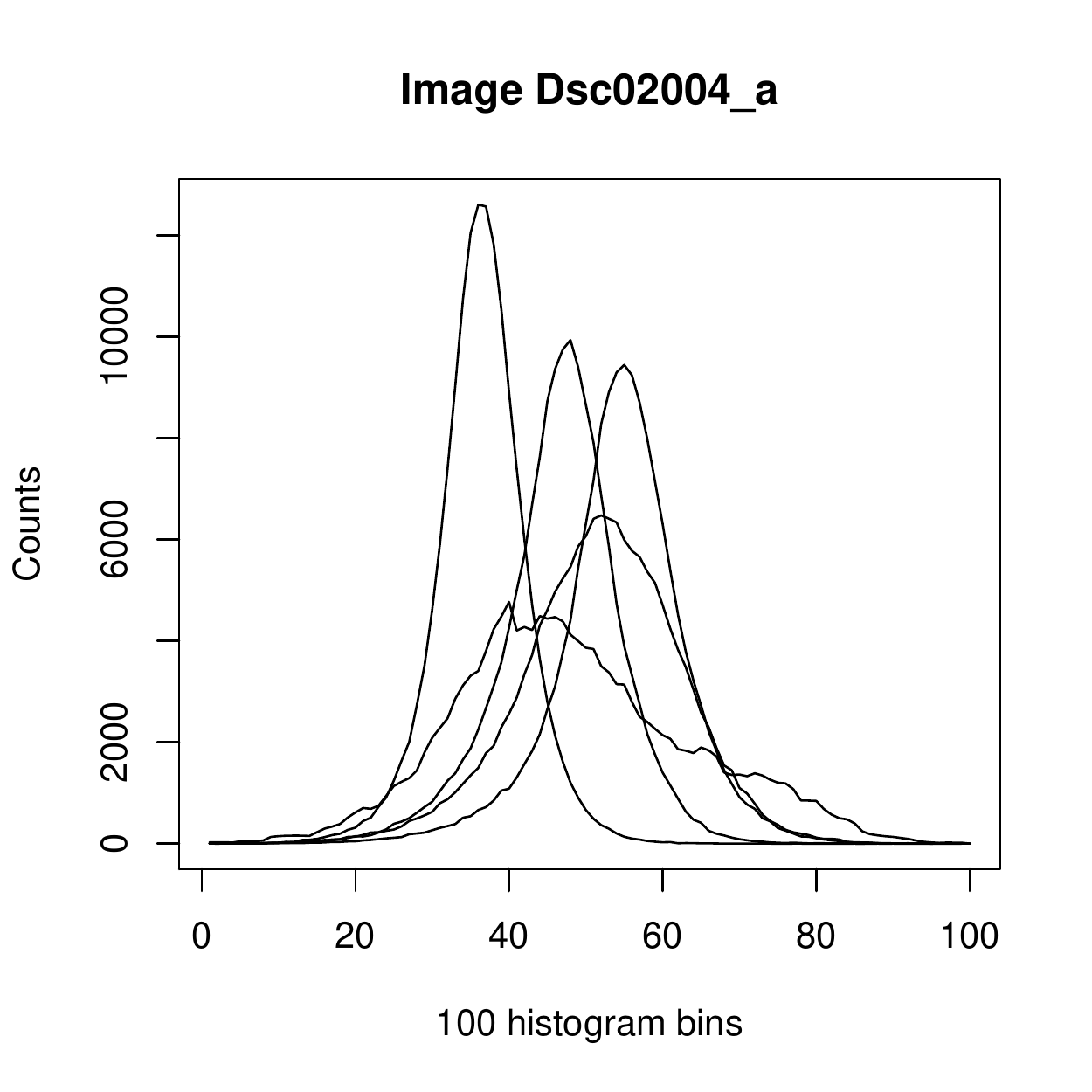} 
   \caption{Distributions of wavelet scales of the image shown in Figure 
\ref{fig1}.  The higher peaked curves are from wavelet scales 1, 3, and 2, 
respectively,
from left to right.  The second lowest peaked curve, and the lowest one, are 
from wavelet scales 4 and 5, respectively.}
   \label{fig2}
\end{figure}

The image shown in Figure \ref{fig1} is from the 
application which motivated this work.   
It is characterized by textured signal, and the image's distributional 
model may be useful 
to us for handling irregular variation in texture.  The distributions of
wavelet scales 1, 2, 3, 4 and 5, furnished by a B$_3$ spline \`a trous 
wavelet transform, were determined using, in each case, a histogram with 
100 bins.  Figure \ref{fig2} shows these distributions.   Wavelet scales 
1, 2 and 5 look somewhat long tailed; on the other hand wavelet scales 3 and
4 look more symmetric.  

We tested the distributions shown in Figure \ref{fig2} (Markwardt, 2004), 
using the
 mean square error (MSE) of the fits by a Lorentzian (also called Cauchy,
a long tailed distribution) 
and a Gaussian to the wavelet scales.  Results are shown in 
Table \ref{tablorgauss}.
We conclude that both Gaussian and Lorentzian distributions may well be 
relevant to practical image analysis situations of the sort considered.  

\begin{table}
\caption{MSEs of fits.  For scales 1, 2, and 5, the fit by a Lorentzian 
outperforms the fit by a Gaussian.  However, for scales 3 and 4, a 
Gaussian fit is better.}
\begin{center}
\begin{tabular}{crr} \hline
Wavelet scale     &    Lorentzian fit   &   Gaussian fit  \\ \hline
1                 &    1.24             &   4.90          \\
2                 &    0.07             &  29.55          \\
3                 &  1411.76            & 594.46          \\
4                 &  73890.6            & 50390.1         \\
5                 &  156515.0           & 271948.0         \\ \hline
\end{tabular}
\end{center}
\label{tablorgauss}
\end{table}

Taking the 6 images to be discussed below (Figure \ref{clim}), 
we looked at the fits
of Lorentzian and Gaussian distributions.  
Table \ref{tab1} shows the results.  For 
each of the selected images we look at Lorentzian versus Gaussian peak fits
to 5 wavelet coefficient levels.  Sometimes the long tailed Lorentzian gives
a better result, maybe even a spectacularly better result.  However this 
is not always the case.  Sometimes the Gaussian fit is far better, and 
we also note that high MSE values may indicate that neither distribution
is particularly good. 

\begin{table}
\caption{Image sequence number chosen: these are the images shown 
(in succession, from upper left) in Figure \ref{clim}. For each image,
5 wavelet resolution scales are studied. 2D Lorentzian and
Gaussian fits are shown: MSE (mean square error) used.  An asterisk indicates
whether Lorentzian or Gaussian fit is better.}
\begin{center}
\begin{tabular}{rccrcr} \hline
Seq. no. & Scale & & Lorentzian & & Gaussian \\ \hline
1   & 1 & * &    31.9  &  &      43.3  \\
    & 2 &   &  1397.2  &  * &      9.1 \\
    & 3 &  * &  2982.0  &  &  10404.7 \\
    & 4 &  * & 77135.4  &  & 122607.0 \\
    & 5 &  * & 192195.0  &  & 276682.0 \\
60  & 1 &    &    37.6  & * &      28.7 \\
    & 2 &  * &   18.7  &   &   134.8 \\
    & 3 &  * & 22180.5  &  &  26668.1 \\
    & 4 &  * &  37069.2  &  &  44615.1 \\
    & 5 &  * &  859.6  &   &   875.7 \\
120 & 1 &  * &    3.3  &   &     5.6 \\
    & 2 &  * &    2.7  &   &     8.1 \\
    & 3 &  * &   23.8  &   &   214.8 \\
    & 4 &    &    2.0  &  * &      0.0 \\
    & 5 &    & 86422.3  & * &       1.4 \\
180 & 1 &    &   49.1  &  * &      6.6 \\
    & 2 &  * &    0.6  &   &     5.4 \\
    & 3 &    &  9817.3  &  * &      74.0 \\
    & 4 &    & 7739.2  &   * &      5.5 \\
    & 5 &  * & 51196.0  &  &   75436.2 \\
240 & 1 &  * &    0.5  &   &     0.8 \\
    & 2 &  * &    0.3  &   &    23.4 \\
    & 3 &    &   88.0  &  * &       5.8 \\
    & 4 &  * &  591.3  &  &   46947.3 \\
    & 5 &  * & 3315.3  &  &  85459.2 \\
300 & 1 &  * &    3.8  &  &     12.2 \\
    & 2 &    & 2506.9  &  * &      10.3 \\
    & 3 &    & 39793.6  & * &       48.3 \\
    & 4 &    & 13137.1  & * &      108.6 \\
    & 5 &  * & 211860.0  &  & 243913.0 \\ \hline
\end{tabular}
\label{tab1}
\end{center}
\end{table}

\subsection{Role of Multiresolution Coefficient Moments}

Wavelet coefficients are often of long tailed (hence 
not Gaussian) distribution but we find them also sometimes to be close 
to Gaussian.  The Shannon entropy is appropriate for  
Gaussian distributed data, whereas the Tsallis non-extensive entropy is 
associated with power law, long tailed data.  In practice, 
distribution mixtures, at different wavelet resolution scales, of long tailed
and Gaussian distributed data must be handled.  

We find  that both Gaussian/Shannon and long-tailed/Tsallis perspectives
are potentially useful in practice.  While we can proxy the former with 
the second order moment (Starck et al., 1998b, Starck and Murtagh, 2006),
the situation is not the same for the latter because of the many possible
distributions, and the importance of higher order moments.  (A Gaussian
is completely characterized by moments of order 1 and 2.) 

For general analysis where 
multiresolution coefficients follow a mixture of distribution families, 
a convenient and practical way to carry out the analysis is by using 
higher order moments of the multiresolution coefficients as proxies of
the unknown, underlying distributions.  
In the absence of a clear distribution holding for the multiresolution
scales, we are better off keeping to 
multiresolution coefficient moments for reasons of effectiveness, 
practicality and convenience.

\subsection{Third and Fourth Order Moments as Features}

The use of higher order moments, beyond the first and second, for 
texture analysis is well-established (Tsatsanis and Giannakis, 1992; 
Chandran et al., 1997; Avil\'es-Cruz et al., 2005).  
Spatial modeling is  used in Popovici and 
Thiran (2004).  We use spatial models as part and parcel of the 
multiscale transforms.  Other, less typical, applications of 
texture analysis using higher order moments include Kim and 
Strauss (1998), who apply this approach to the ``textures'' of 
point pattern distributions.  

We have motivated higher order moments in the context of 
long tailed distributions of multiscale transform coefficients.  
In the next section  we will illustrate experimentally the potential 
usefulness of higher order moments in the multiscale transform 
context.

\section{Selection of Multiresolution Features: Setting the Context}
\label{sect3}

\subsection{Wavelet Transform and Curvelet Transform}

With the B$_3$ spline \`a trous redundant 
wavelet transform
(Starck et al., 2007), 
there are no aliasing effects due to decimation, and the 
wavelet function (similar to a Mexican hat function) is symmetric, and 
within the limits of separability of use in horizontal and vertical image
directions it approximates an isotropic function.  See Starck et al.\ (1998b),
Starck and Murtagh (2002), and Starck et al.\ (2006).  The (pixelwise 
additive) decomposition of the image was, for all experiments described 
below, 5 wavelet resolution scales plus the smooth continuum image.  

This wavelet transform uses a particular set of basis functions, which are
defined by roughly isotropic functions present at all scales and locations.
Hence this wavelet transform is appropriate for isotropic features or 
mildly anisotropic features.  
To move beyond the wavelet transform, a 
range of other basis function sets have been used, with properties relating
to alignments, elongations, edges, and indeed 
curved features.  Non-wavelet multiresolution transforms
therefore target the detection and characterization of non-Gaussian 
signatures in the image data.  

In Starck et al.\ (2004, 2005) the kurtosis (fourth order moment) was used
to understand the nature of complex non-isotropic features in cosmology.  
The skewness 
and variance (third, second order moments) were also discussed.  
Consequently we wished to investigate the use of these moments in our work.  

The ridgelet transform uses wavelet-like functions which 
are constant along lines $x_1 \cos \theta + x_2 \sin \theta = $ Const., 
where a fixed set of angles $\theta$ is used; and $x_1, x_2$ are related to 
scaling through a dyadic multiplicity factor.  The ridgelet transform can
be shown to be the application of a 1-dimensional wavelet transform to 
constant angle slices of the Radon transform.   The ridgelet transform 
is a good pattern matcher for sheets at varying scales and positions, whereas
the redundant \`a trous wavelet transform targets (isotropic or 
near isotropic) clusters.  

To 
find curved features the curvelet transform is used.  
The curvelet transform first decomposes the image into a set of wavelet
bands.  Then each band is analyzed with a local (i.e., blockwise) ridgelet
transform.  See Starck et al.\ (2002).   The curvelet transform is an 
effective tool for curve finding at multiple resolution levels. 
The command  {\tt cur\_stat} in the MR package (MR, 2004) was used
for the curvelet transform.  Six scales 
were used, with a ridgelet block size 
of 16.  This gave a total of 19 curvelet coefficient bands.  


\subsection{Data}

In this section we will present, using a simple procedure, how we can 
show that (i) multiscale transforms other than wavelet transforms,
and (ii) higher order moments, may provide the most discriminating features, 
This is a ``proof of concept'' demonstration, based on a simple but 
non-trivial image classification problem.  

We took four images with a good quantity of curved edge-like structure
for two reasons: firstly, due to a similar mix of smooth, but noisy 
in appearance, and edge-like regions in our construction images; and 
secondly, in order to test the curvelet as well as the wavelet transforms.
To each image we added three realizations of  Gaussian noise of standard 
deviation 10, and three realizations of Gaussian noise of standard 
deviation 20.  Thus for each of our four images, we had seven realizations
of it.  In all, we used these 28 images.   

\begin{figure}[htbp] 
   \centering
   \includegraphics[width=2.5cm]{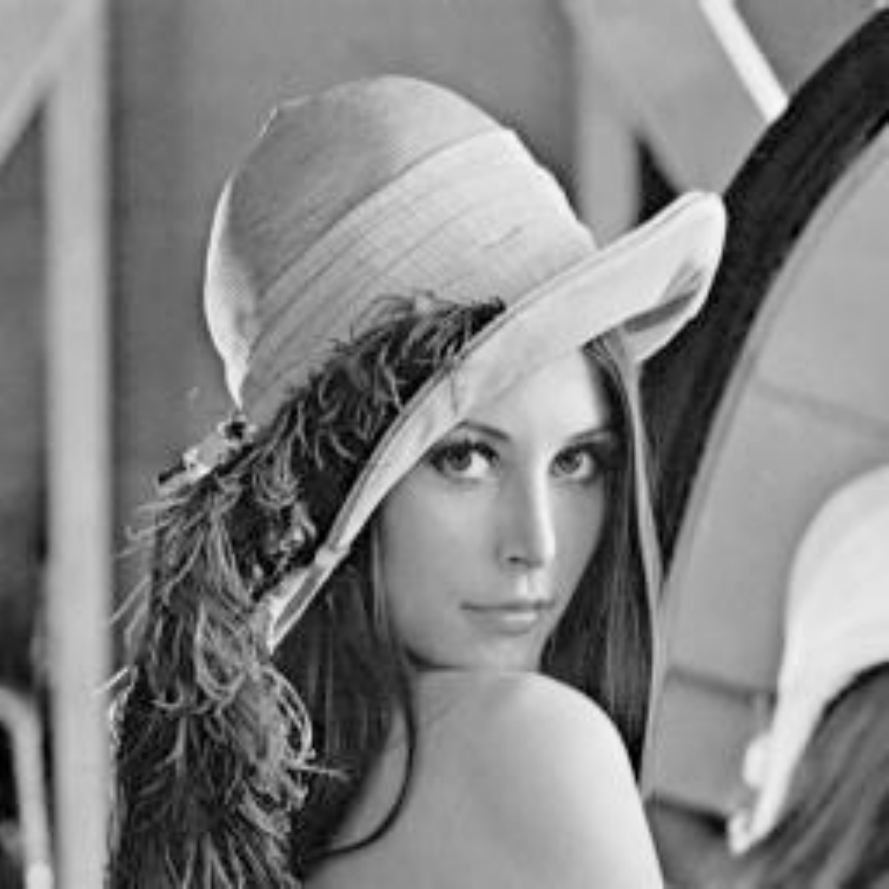} 
   \includegraphics[width=2.5cm]{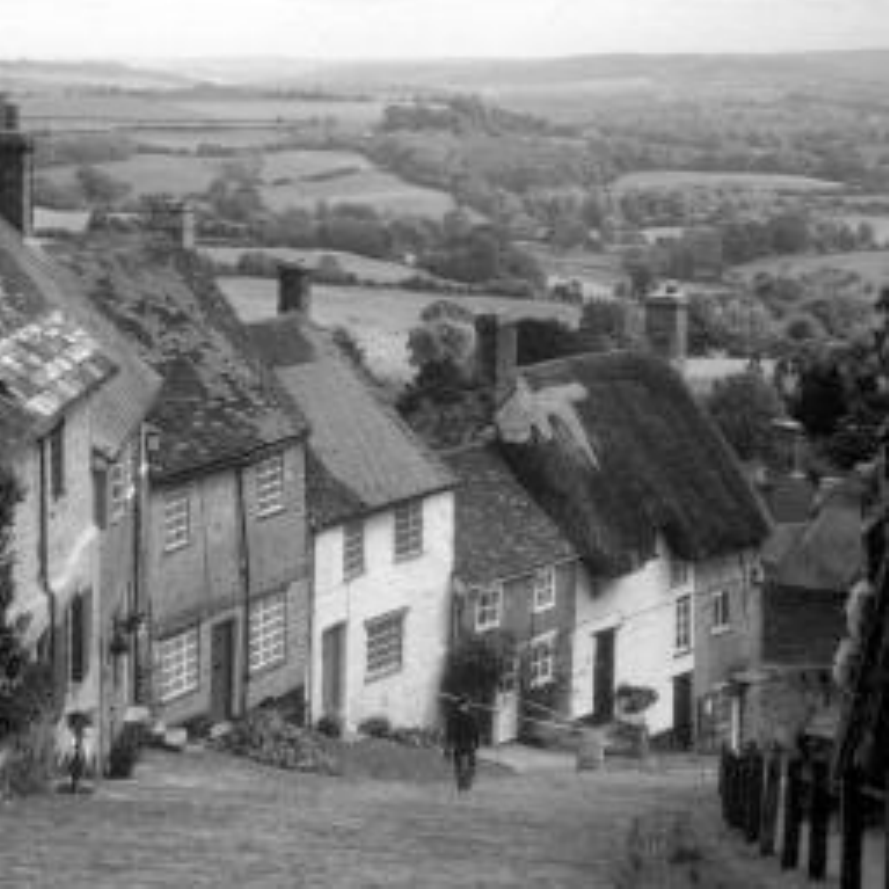} 
   \includegraphics[width=2.5cm]{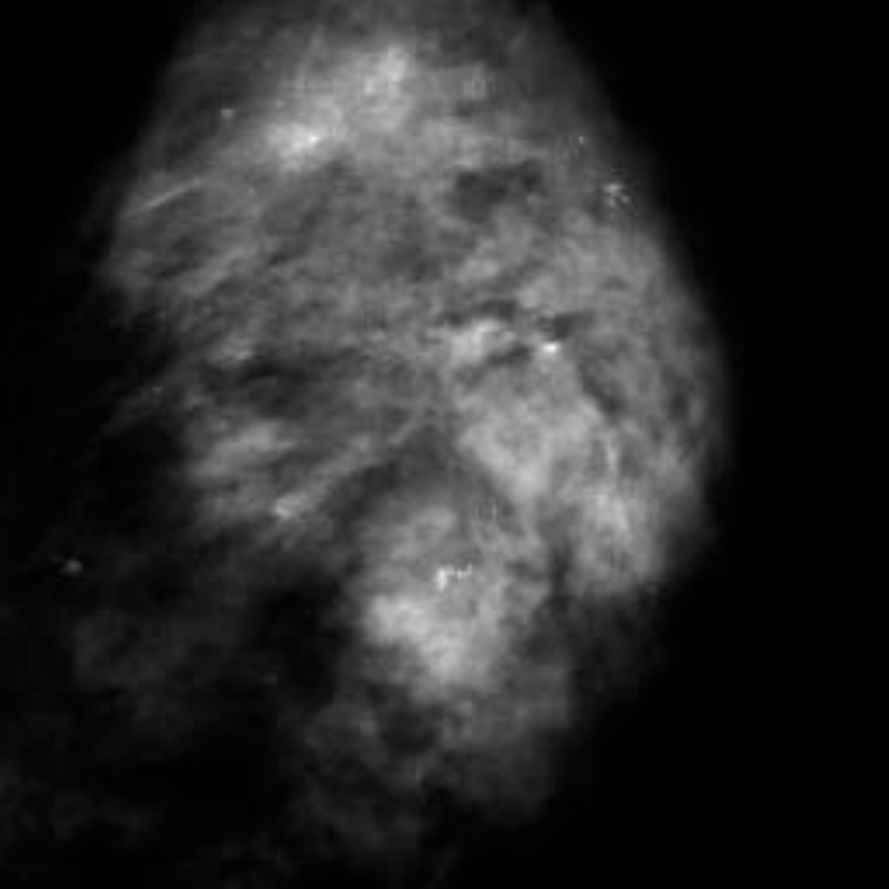} 
   \includegraphics[width=2.5cm]{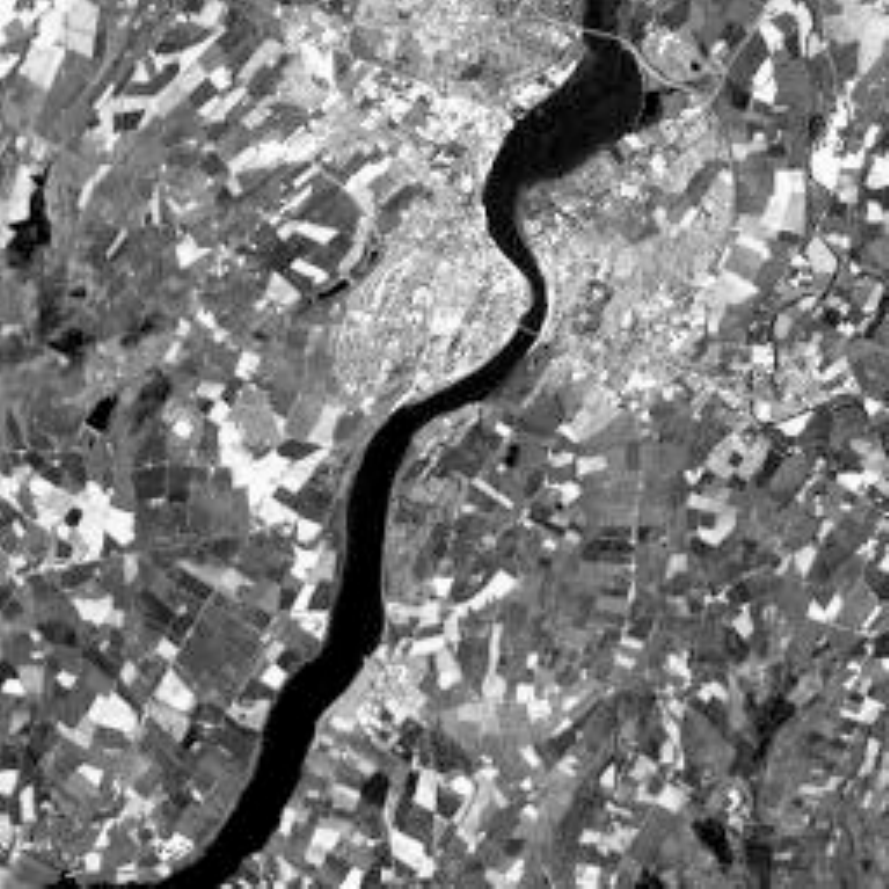} 
   \includegraphics[width=2.5cm]{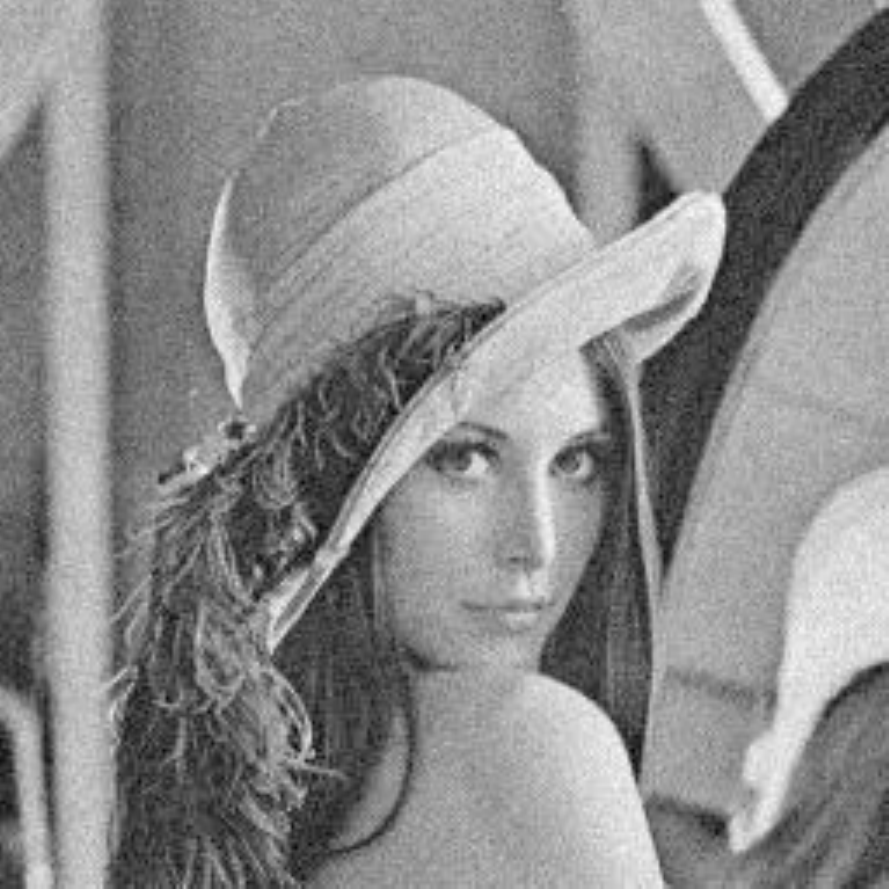} 
   \includegraphics[width=2.5cm]{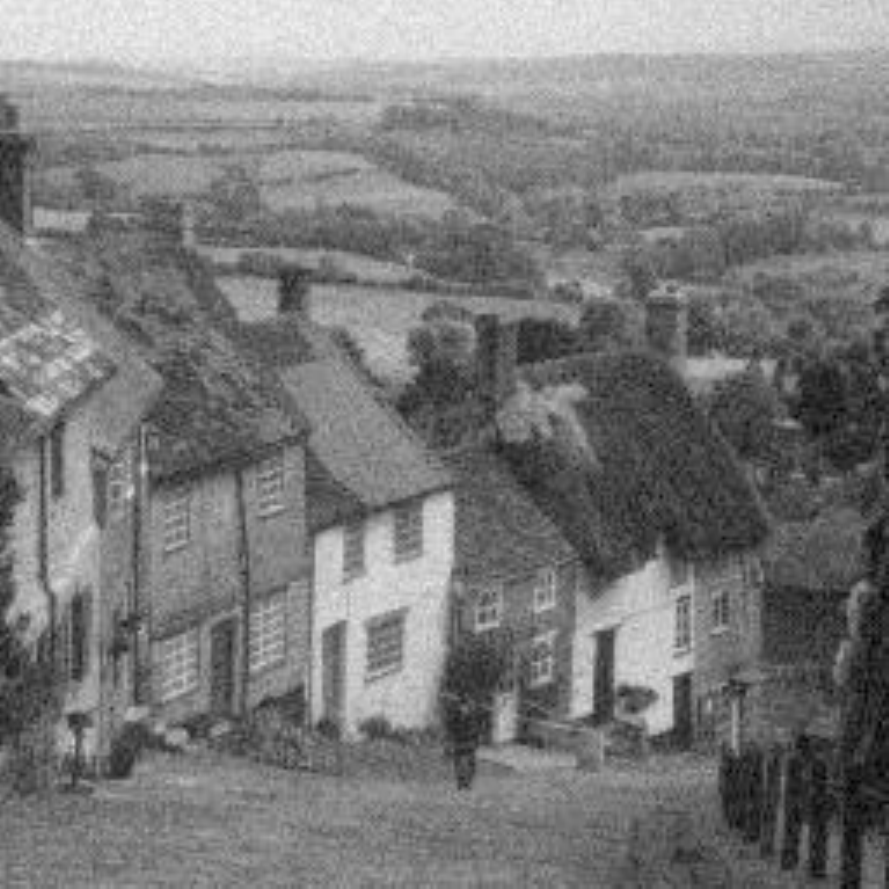} 
   \includegraphics[width=2.5cm]{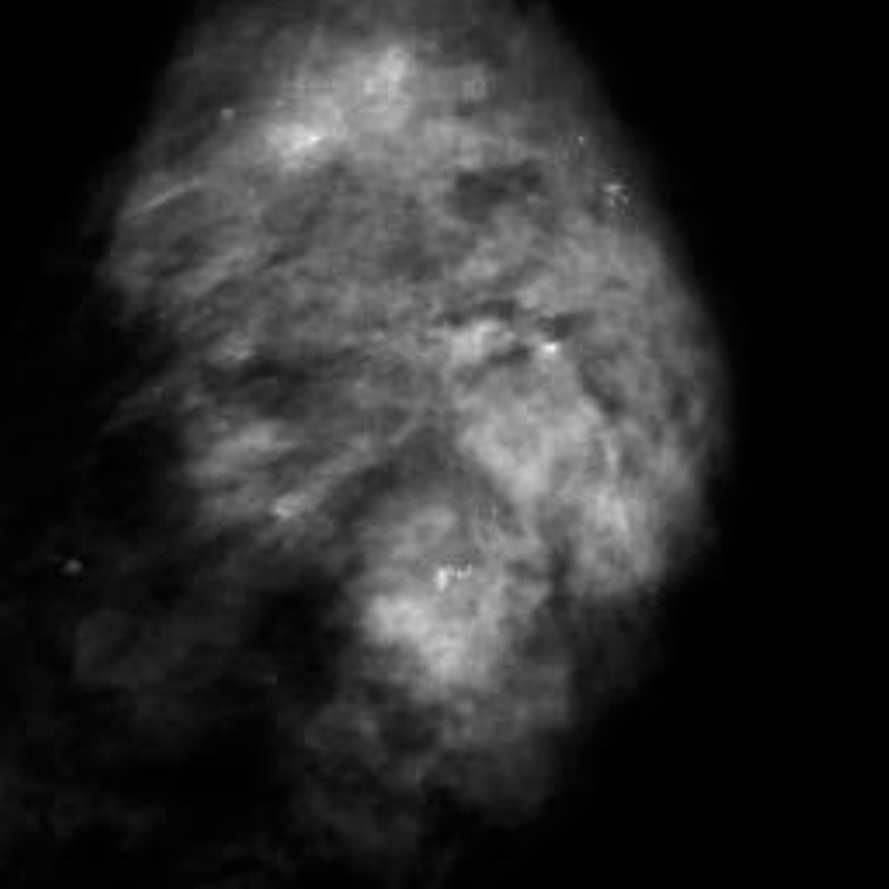} 
   \includegraphics[width=2.5cm]{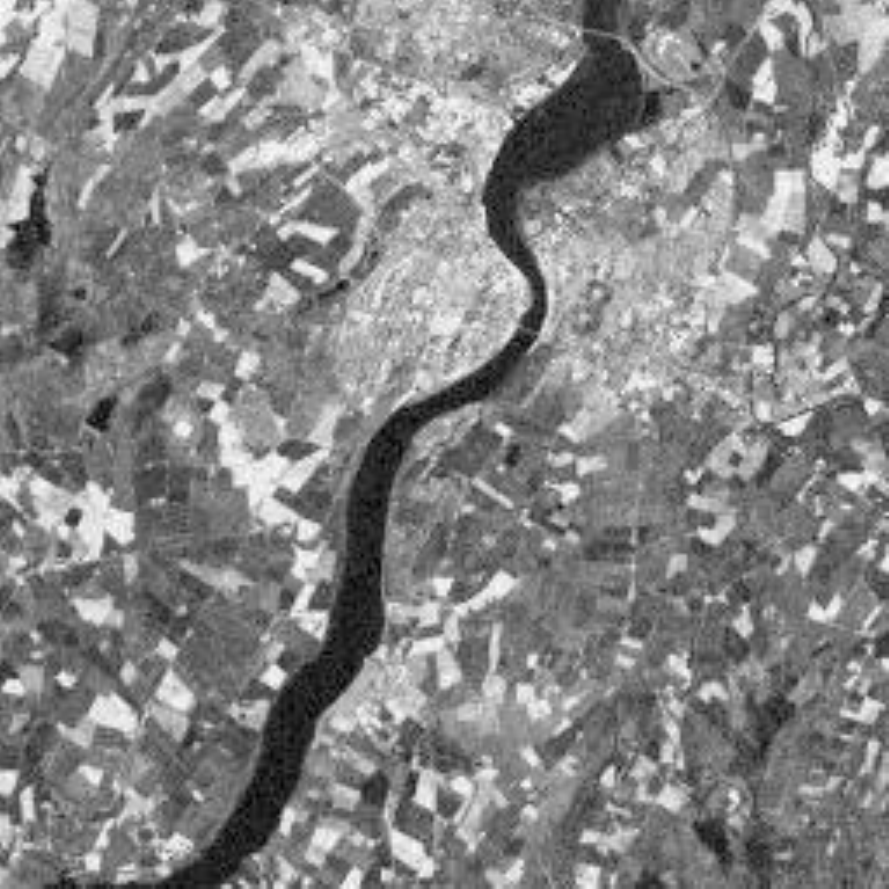} 
   \caption{Four images used (top row), and (bottom row)
each with added Gaussian noise of standard deviation 20.}
   \label{clim0}
\end{figure}

Examples are shown in Figure \ref{clim0}.  The images used were
all of dimensions $512 \times 512$.  The images were the widely used 
test images Lena and Landscape, a mammogram, and a satellite view of the 
city of Derry (Londonderry) and River Foyle in Northern Ireland.  

We expect the effect of the added noise to make the image increasingly smooth 
at the more low (i.e., smooth) levels in the multiresolution transform.  

The data used therefore was the set of multiresolution transform coefficients
for each of the 7 images relating to one of our four test images.  What 
we will seek to do is to find very clear similarity between the 7 images 
that are all derived from one initial image.  So we will seek a very clear
discrimination between the four clusters of image, each cluster having 
7 images.  

Our analysis aims at distinguishing between clusters of images, and 
determining the most useful features for this.  We address these analyses
in an integrated way, taking all data into account simultaneously.

The 28 images are each characterized by: 

\begin{itemize}
\item For each of 5 wavelet scales 
resulting from the \`a trous wavelet transform, we determined the 2nd, 
3rd and 4th order moments at each scale (hence: variance, skewness and
kurtosis).  So each image had 15 features.
\item For each of 19 bands resulting from the curvelet transform, we again
determined the 2nd,
3rd and 4th order moments at each band (hence: variance, skewness and
kurtosis).  So each image had 57 features.
\end{itemize}

The 28 images were therefore characterized by 72 features, taking 
spatial and frequency resolution scale into account.  We did not 
normalize the images, notwithstanding the varying pixel means.  This 
decision was made in order to avoid the choice of any  
ad hoc way of doing this.
For the analysis of feature importance, and of how well we can cluster 
our 28 images into four clusters, an important requirement ensues: we 
must only use relative values, or what we can term a {\em profile} of values,
and not the absolute values.  

This issue of relative values is very adroitly 
handled by correspondence analysis
(Murtagh, 2005).  Just as with principal components analysis, the 
inherent dimensionality of both our 28 images in a 72-dimensional space,
and our 72 features in a 28-dimensional space, must be the minimum of 
28 and 72.  Call the value on feature $j$ for image $i$ $x_{ij}$, and 
convert it to a fraction bounded by 0 and 1: $f_{ij} = x_{ij}/x$ where 
$x = \sum_i \sum_j x_{ij}$. 
Correspondence analysis forms profiles both by row (image) and by column
(feature): $x_{ij}/x_j$ for each row, where $x_j$ is the column sum; 
and $x_{ij}/x_i$ for each column, where $x_i$ is the row sum.  
Assume $f_{ij} \geq 0$.  Given the Gaussian noise, this was not always the
case: our sole modification was to enforce a mass, $f_i$ or $f_j$ to be 
$\geq 0$. 

The $\chi^2$
metric is used, defined for two rows $i$ and $i'$ as: 
$\sum_j 1/f_j (f_{ij}/f_i - f_{i'j}/f_{i'})^2$.  A new set of coordinate
axes are found to best fit the data in both feature (28-dimensional) and
image (72-dimensional) spaces.  This output {\em factor} space is endowed with 
the Euclidean metric, allowing visualization.  Unlike principal components
analysis, the scales of both feature and image spaces are the same, so that
both rows and columns can be displayed in the output representation.

The percentage inertia explained by the first factor (tantamount to the 
overall information content explained by this factor) was 86.9\%, 
indicating a highly one-dimensional underlying manifold in the dual
spaces of images and of features.  Note again that while the absolute
input values varied greatly in accordance with originating image, the
use of {\em profiles} in correspondence analysis guarantees that this 
very pronounced one-dimensionality is a characteristic of the data.  

\begin{figure}
\centering
\includegraphics[width=14cm]{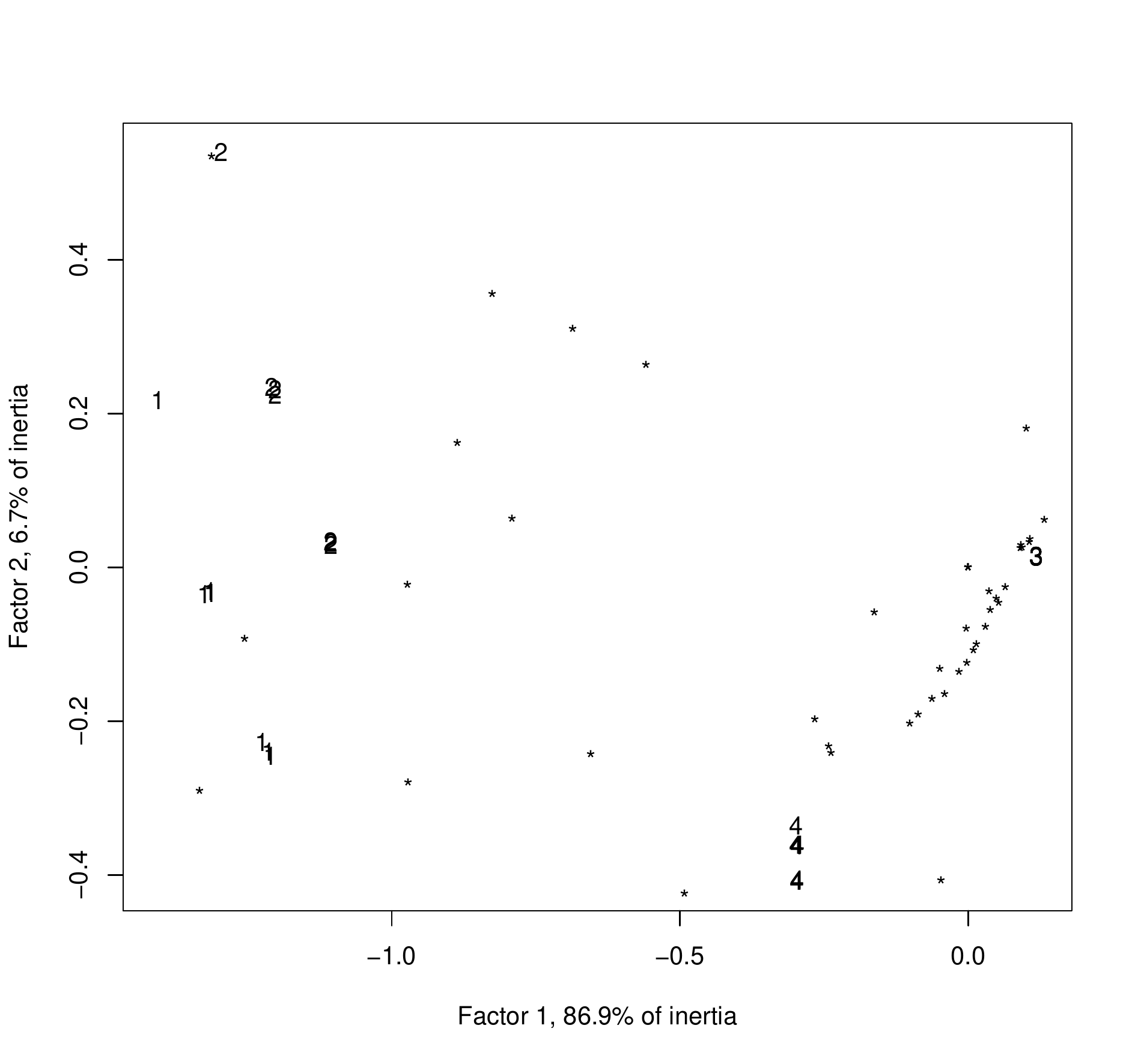}
\caption{Principal factor plane.  Clusters of images are displayed with 
1 for first cluster images, 2 for second, and so on.  Features are 
displayed with an asterisk.}
\label{ca1}
\end{figure}

Figure \ref{ca1} shows both images and features on the principal factor 
plane.  The images in cluster 3 (the mammogram ones) are completely 
superimposed.  The images in cluster 4 (the Derry ones) are close.
The images in the two other clusters (cluster 1: Lena; and cluster 2: 
landscape) are arrayed somewhat diagonally.  In all cases there is clear
distinction between image clusters.  

\begin{figure}
\centering
\includegraphics[width=12cm]{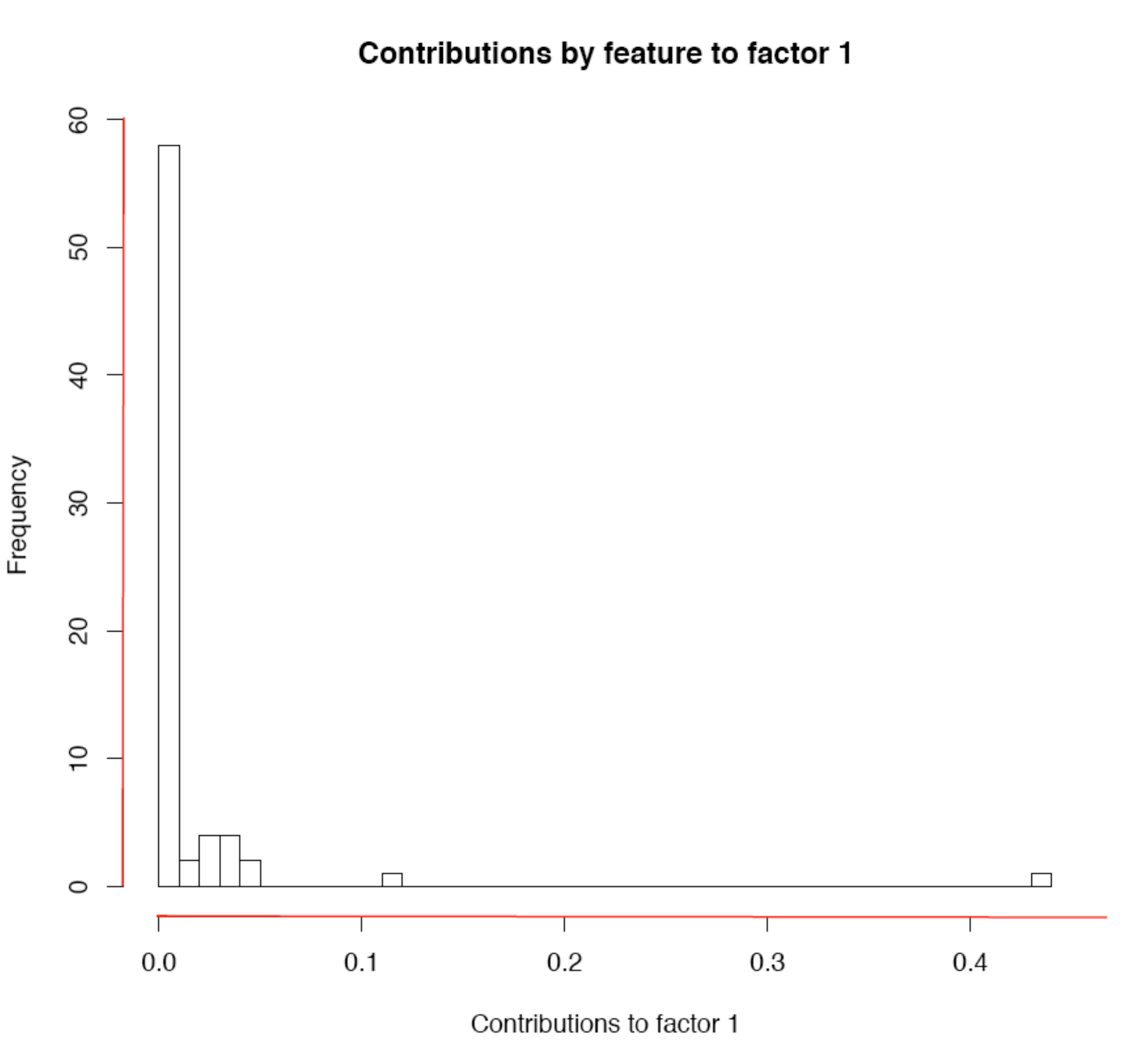}
\caption{Histogram of values of contributions by features (abscissa).
In looking for features with strong contributions to the factor, we 
find just two here, with contributions of 0.116 and 0.437.}
\label{contrhist}
\end{figure}

To assess influence of features, we can avail of the {\em contributions},
defined as sum of mass times projected distance squared, of the features 
relative to the first (predominant)
factor.  For feature $j$, its mass is $f_j$.  Let its projected
value on factor 1 be $F_1(j)$.  Then its contribution to this factor is 
$f_j F_1^2(j)$.  Contributions are commonly used in correspondence analysis
to interpret the results (Murtagh, 2005).  Just two features are found to be 
important.  These correspond to the
two ``blips'' with contribution values 0.116 and 0.437 in
the histogram shown in Figure \ref{contrhist}.  These features relate to the
curvelet transform in both cases.  Firstly band 12 and secondly band 16
are at issue.  In both cases it is a matter of the 4th order moment.

\begin{figure}
\centering 
\includegraphics[width=14cm]{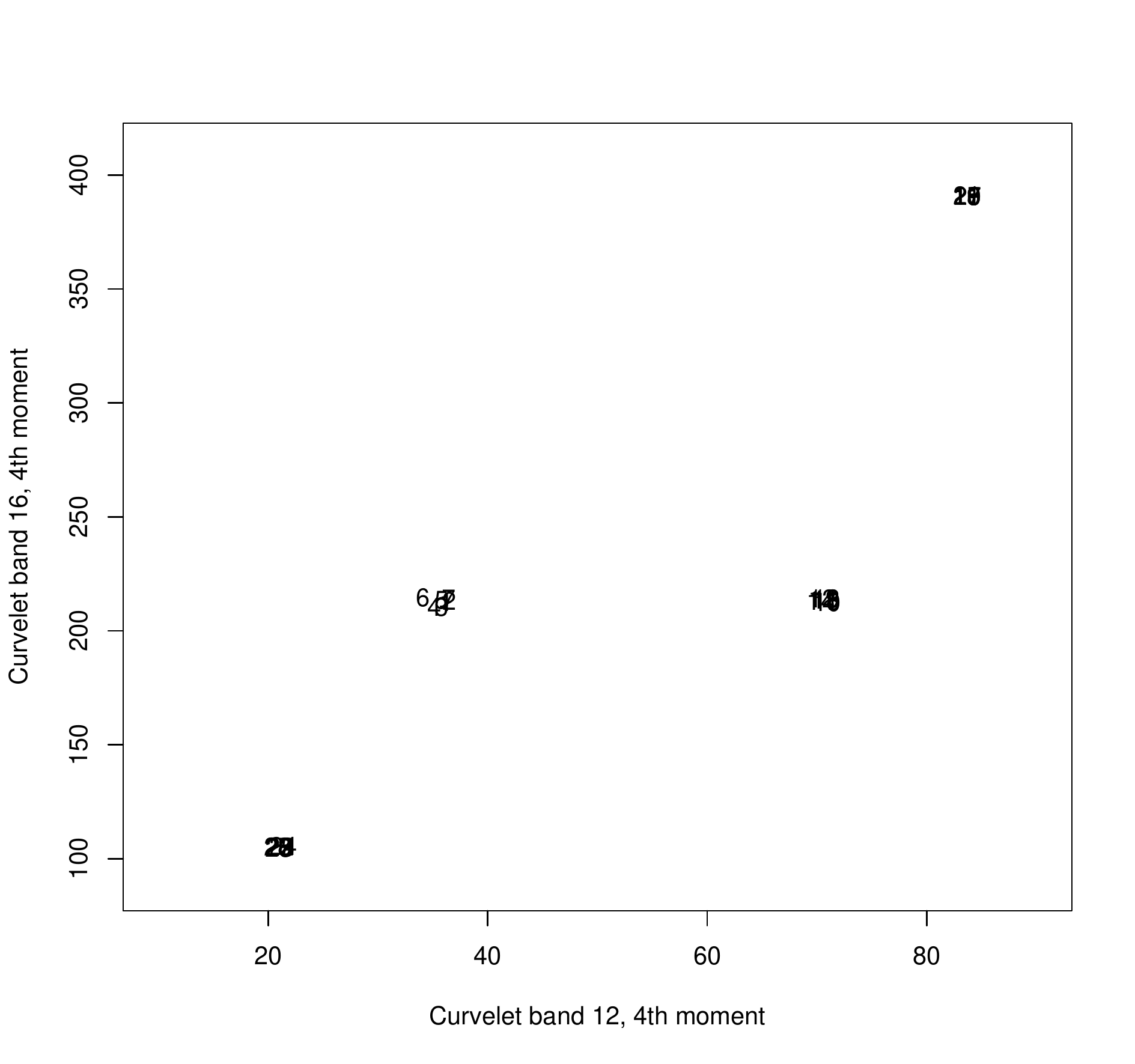}
\caption{Clusters 
of images labeled 1--7, 8--14, 15--21, and 22--28 shown as curvelet
transform band 12 and band 16, 4th order moment (in both cases) 
of coefficients.  See Figure \ref{hicontrib2} for application of jitter
to expose the superimposed points.}
\label{hicontrib1}
\end{figure}

\begin{figure}
\centering
\includegraphics[width=14cm]{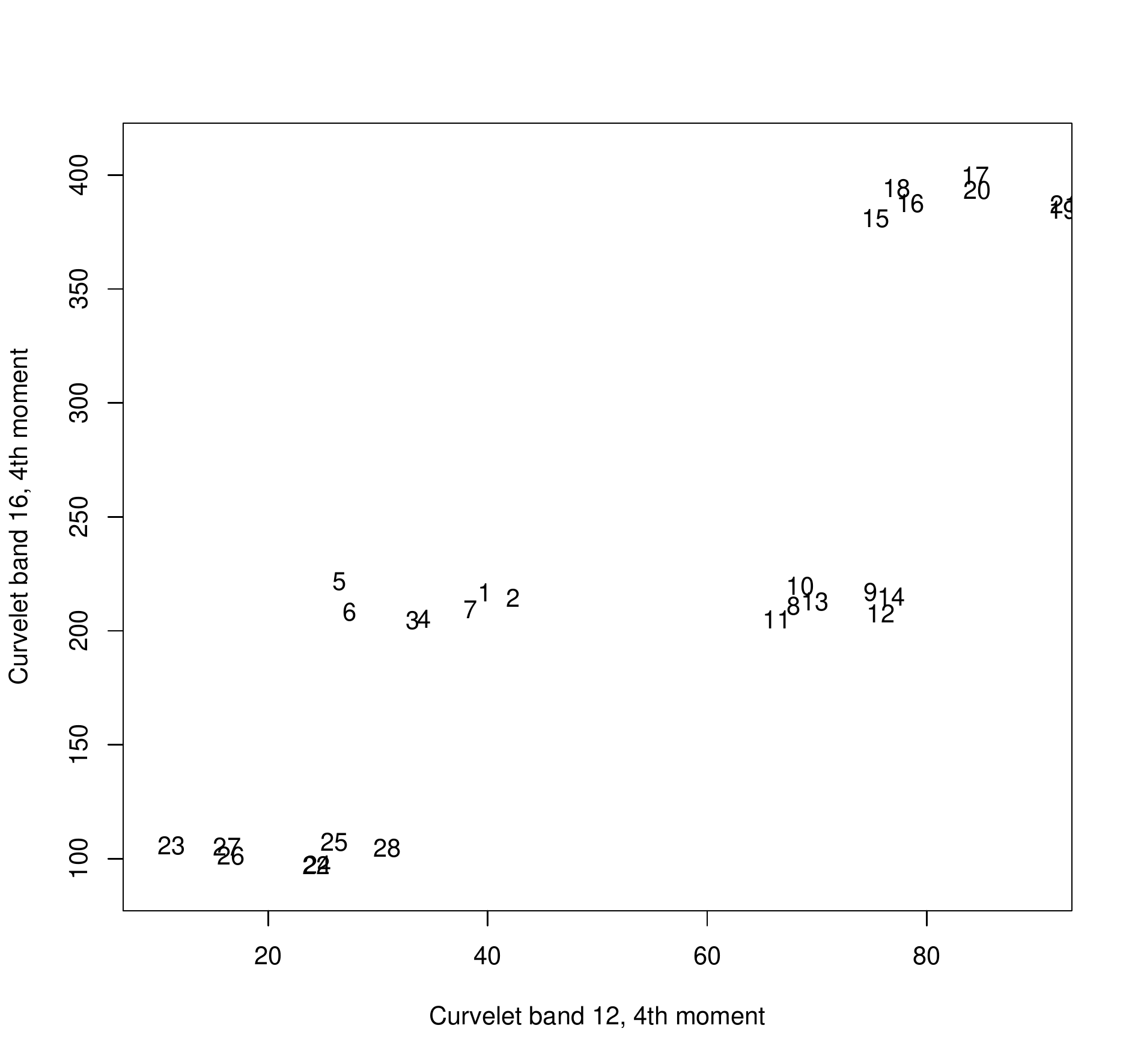}
\caption{Identical to Figure \ref{hicontrib1} but with extensive jitter 
applied to image labels to avoid the superimposed display.} 
\label{hicontrib2}
\end{figure}

That these contributions are so pronounced should manifest itself in 
image cluster low dimension projected 
locations: if anything, the use of these two features alone 
should make our image clusters even more compact.  We see this in Figure
\ref{hicontrib1}.  Many labels of images are superimposed there.  So we 
extensively jittered the points displayed there in Figure \ref{hicontrib2}.
This is an erroneous display but it helps to understand Figure 
\ref{hicontrib1}.

It is obvious that for a given collection of images, some multiresolution 
feature, or set of features, 
may do just the right job in providing a best discrimination.  Our assessment
framework has found that two curvelet, 4th order moments, are far and 
away  the best
for the image collection used.  

\section{Application to Image Grading}
\label{sect4}

\begin{figure}[htbp] 
   \centering
   \includegraphics[width=4cm]{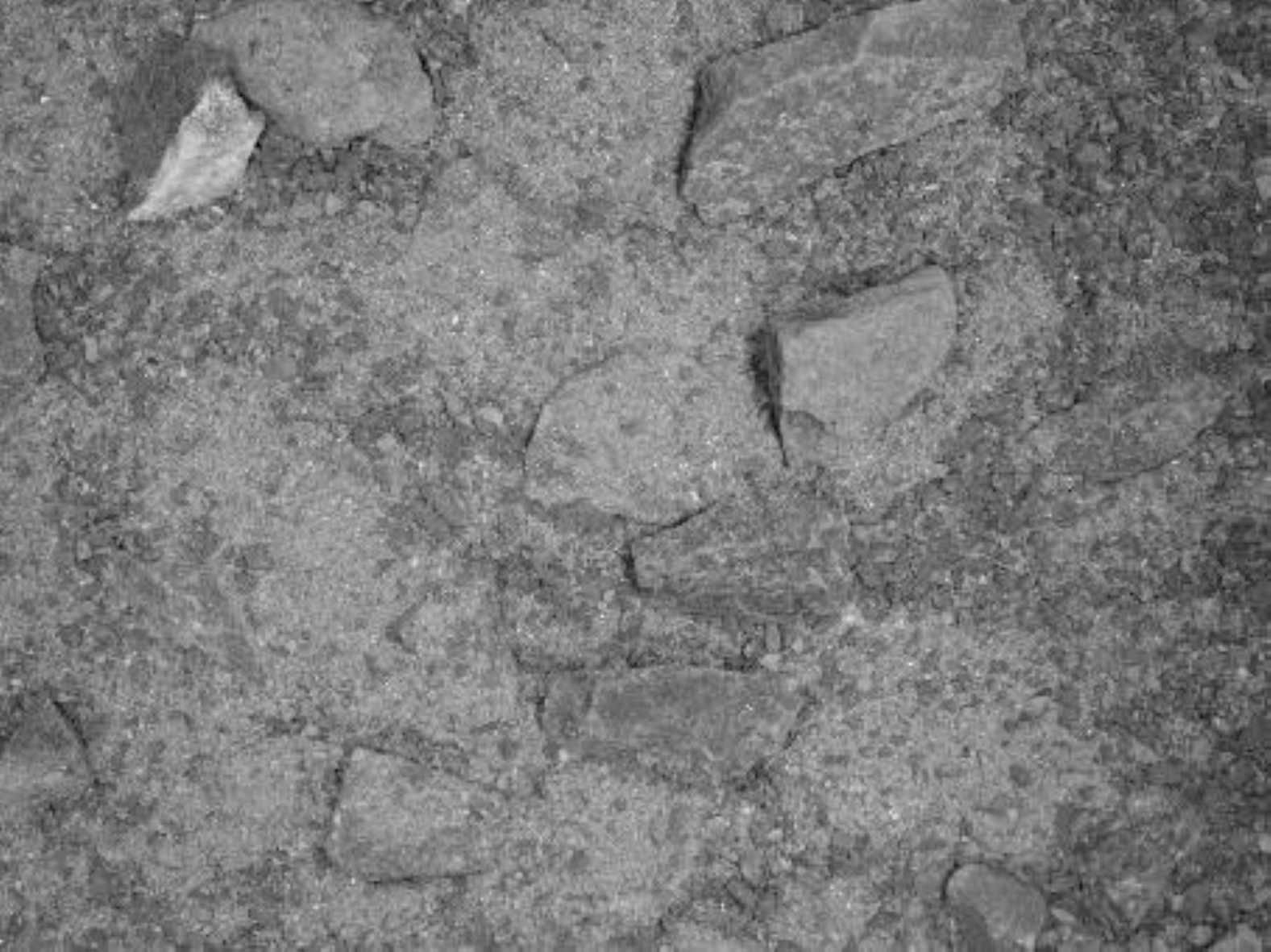} 
   \includegraphics[width=4cm]{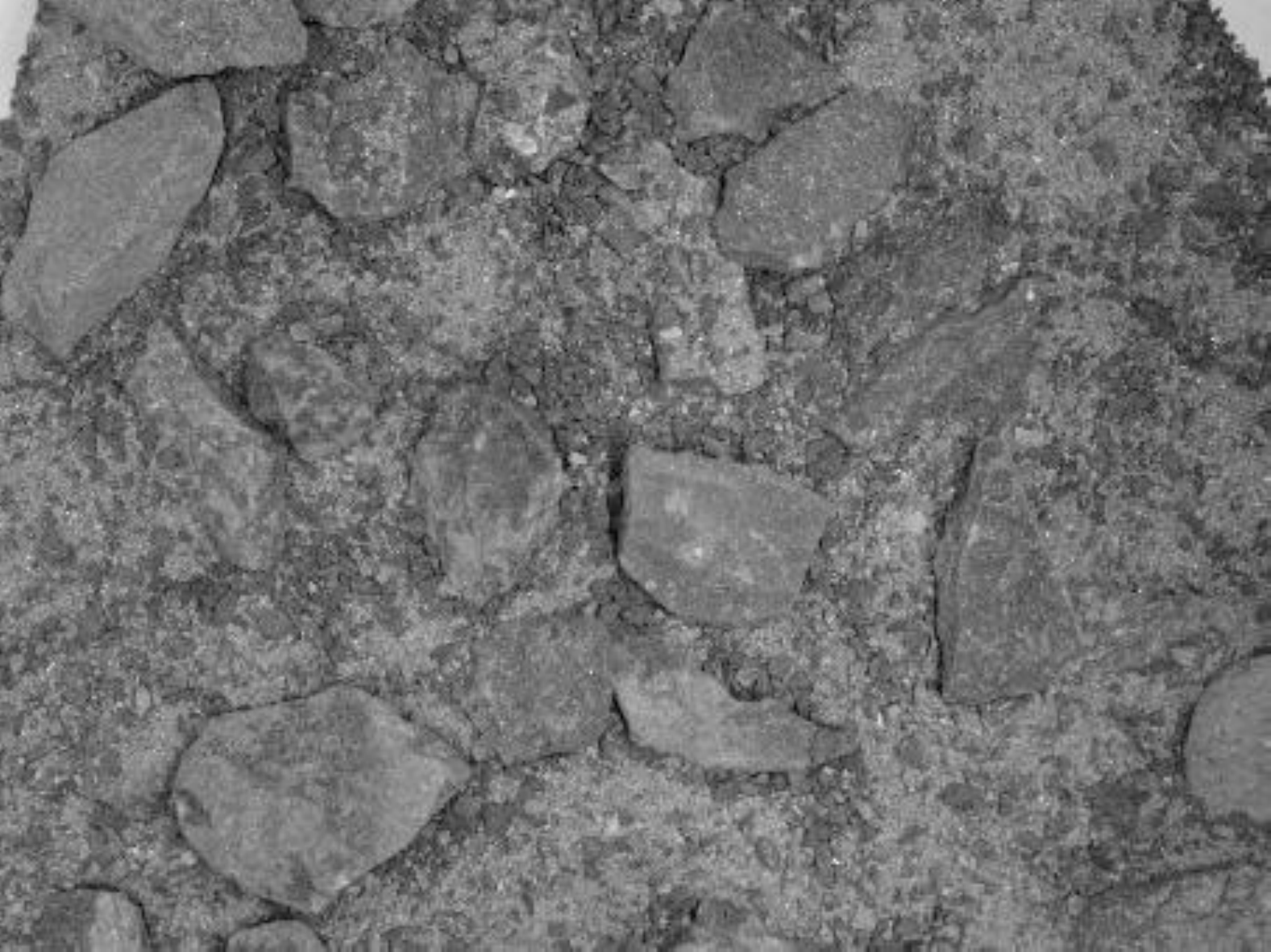} 
   \includegraphics[width=4cm]{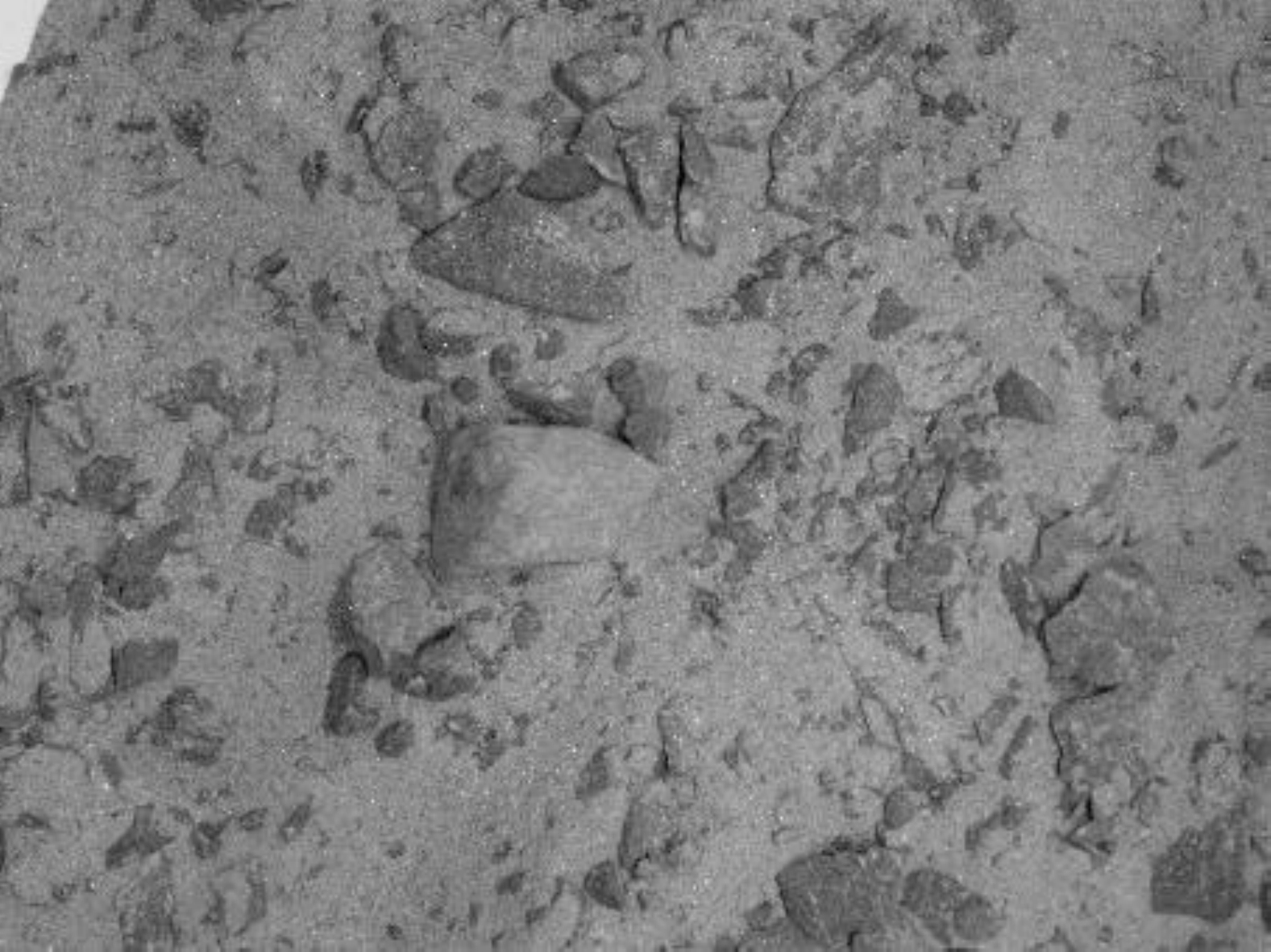} 
   \includegraphics[width=4cm]{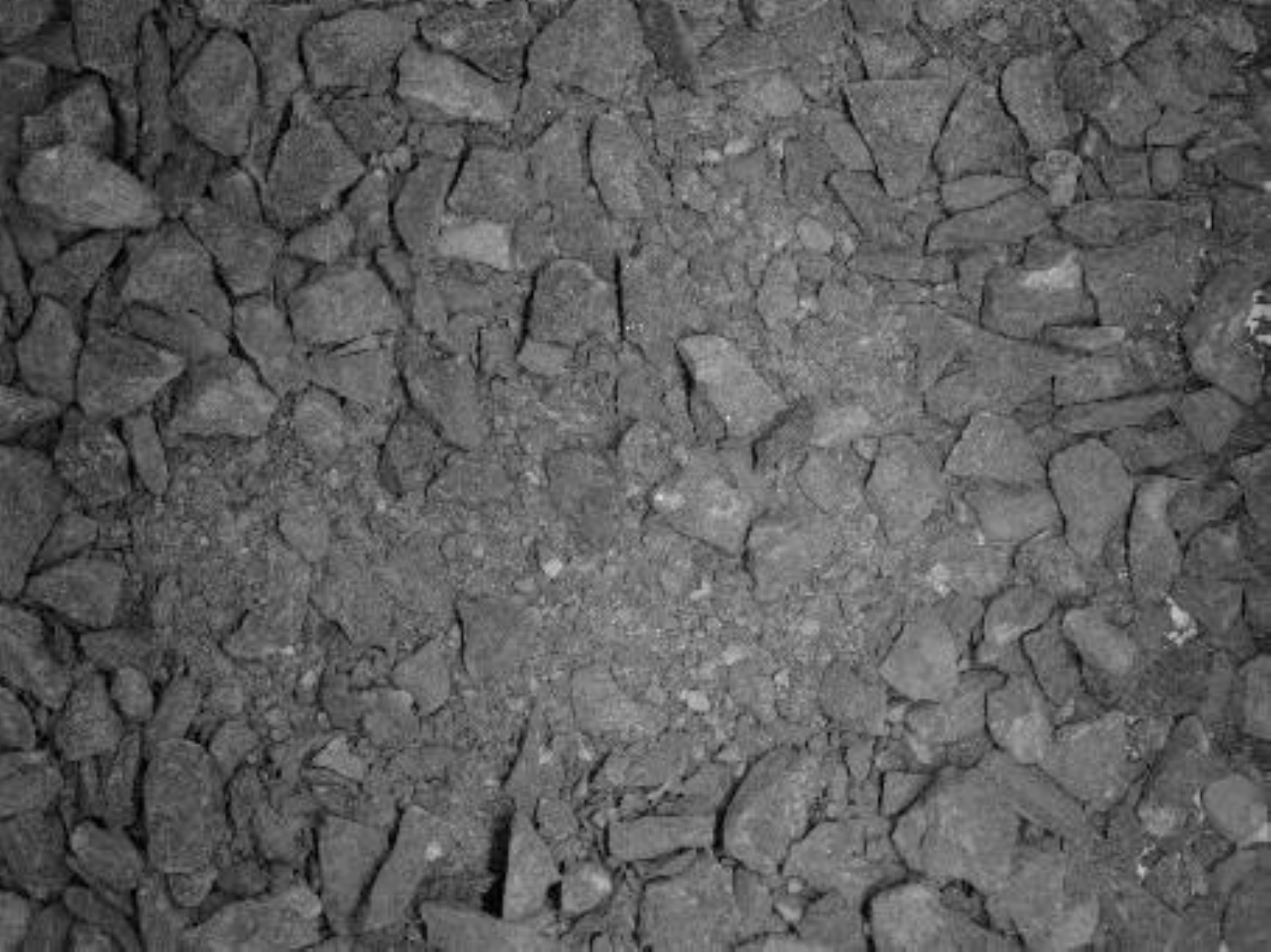} 
   \includegraphics[width=4cm]{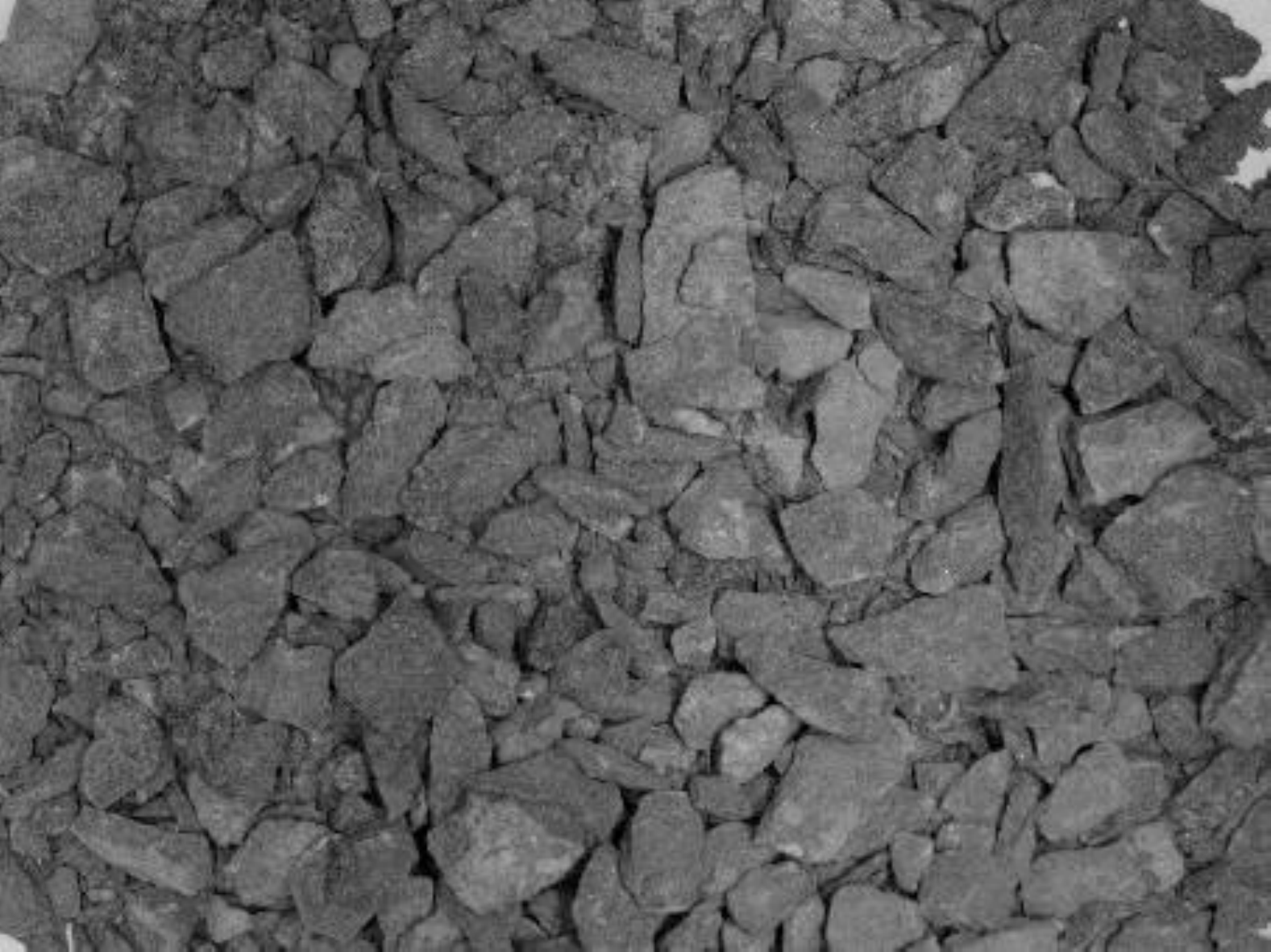} 
   \includegraphics[width=4cm]{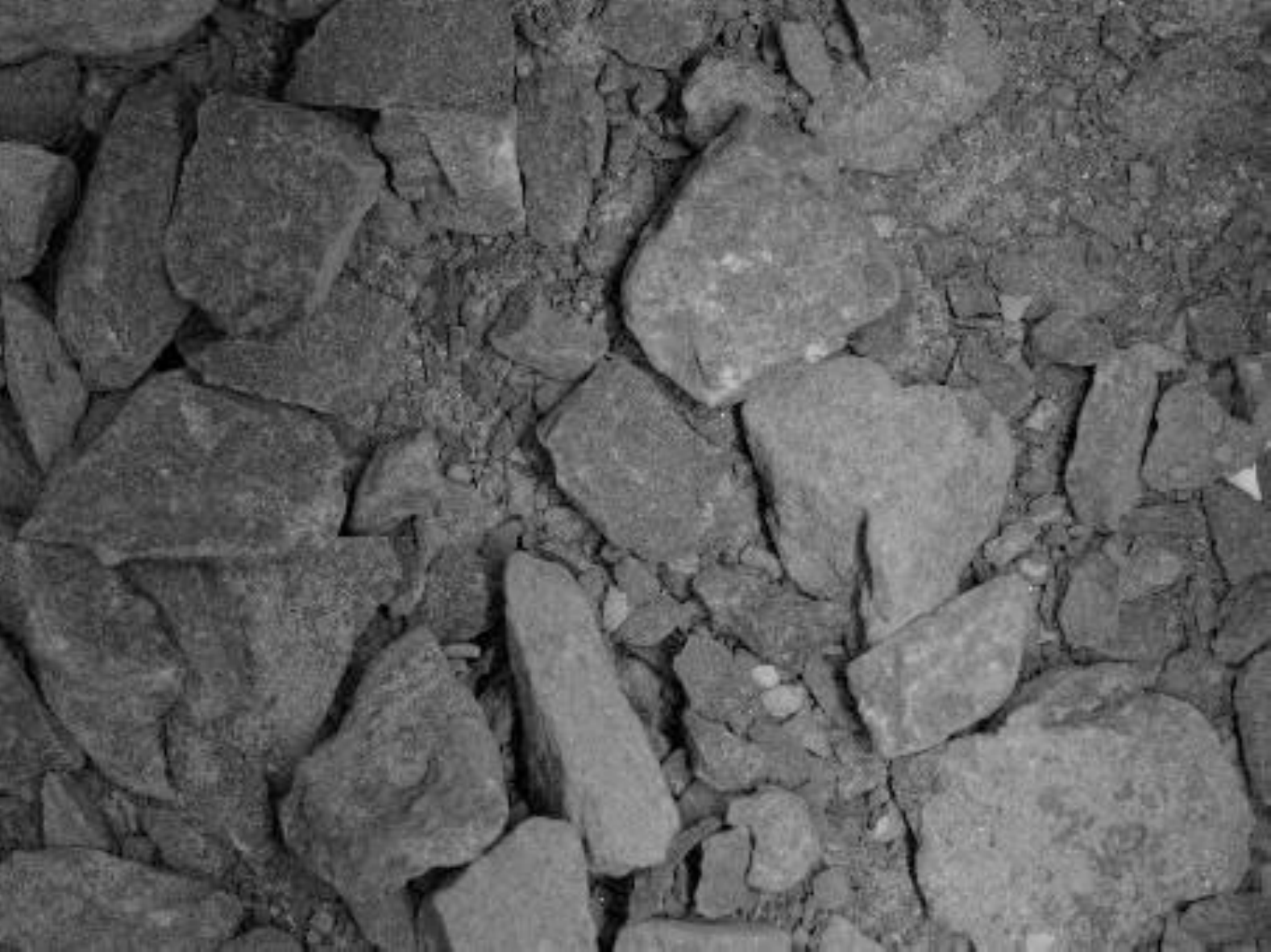} 
   \caption{Sample images from classes 1 through 6, in sequence 
    from upper left.}
   \label{clim}
\end{figure}

The image grading problem related to construction materials
and involving discrimination of aggregate mixes, is exemplified in 
Figure \ref{clim}.  
The data capture conditions included (i) constant height of camera 
above the scene imaged, and (ii) a constant and soft lighting resulting from
two bar lamps, again at fixed height and orientation.  
It may be noted that some of the variables we use, in particular
the variance, would ordinarily require prior image normalization.
This was expressly 
not done in this work on account of the relatively homogeneous 
image data capture conditions.  In an operational environment such 
a standardized image capture context would be used.  

The British Standard specification sets out nominal proportions of constituent
materials in a mix, which we call a class, in terms of sieve size.
Classes, in such constituent property space, are overlapping.
Our first approach was therefore as follows.  With different feature
sets we carried out extensive testing of the discrimination properties,
initially with training sets from class boundary regions, and
testing on images from the central regions of the classes.  But with
overlapping classes of irregular and unknown morphologies in any
constituent property space, this was not a productive approach.
We instead therefore used training images from central regions of the
classes, and test images from class boundaries.  Our training data
consisted of 12 classes of 50 images, and we selected 3 classes
(classes 2, 4 and 9, spanning the 12 classes) each of 100 images as
test data.

As before we used 5 wavelet scales from a B$_3$ spline \`a trous redundant  
wavelet transform, and for each scale we determined the wavelet 
coefficients' variance, kurtosis and skewness.  Similarly, using the 
curvelet transform with 19 bands, for each band we determined the curvelet
coefficients' variance, kurtosis and skewness.  As before we used 
therefore 72 features.  Our training set comprised three classes, each of 
50 images.  Our test set comprised three classes, each of 100 images.  

Our features are diverse in value, and require some form of normalization.
However we know from the discussion in section \ref{sect2} 
that, for instance, reduction to 
unit variance would be inappropriate if the feature values are non-Gaussian.
So we again use a correspondence analysis of all the data available to us, 
a superset of the data so far described, -- in all 12 classes 
(incorporating the 3 classes defining our training data) of 50 images 
each, and the 300  test images.  The correspondence analysis was carried 
out on 900 images, each characterized by 72 features.  One important aim 
was to map the data, both images and features, into a Euclidean space as a
preliminary step prior to using k-nearest neighbors discriminant analysis.
  
The first axis was very dominant (75.7\% of the inertia was explained by it), 
and again a curvelet coefficient feature was the most dominant in 
the definition of this factor: it related to the 16th band in the 
curvelet transform, and (again) the 4th order moment, or skewness.  

The assignments of test data (three classes, called here classes 2, 4 and 9,
each of 100 images) to the training data (these three classes, each of 
50 images) was assessed using k-nearest neighbors.  This supervised 
classification approach was used in view of the 
difficulty level of this data (we looked at low dimensional 
displays resulting from the correspondence analysis) and the nonlinear 
properties provided by k-NN. We used k = 1.  We assessed:

\begin{itemize}
\item The 72-dimensional feature space.
\item The 71-dimensional factor space.  (71, because of a linear dependency
through centering the factor space cloud; if there are $n$ rows and $m$ 
columns, then this Euclidean embedding dimensionality is 
min($n - 1$, $m - 1$)).  
\item Then we explored {\em all} low dimensionality spaces, using the 
ordering of factors.  This would make no sense if we did not have a
coordinate system with an ordering (of ``importance'', provided by the 
percentage inertia explained) of the coordinates. 
\end{itemize}

\begin{table}
\caption{Assignments, to classes labeled 2, 4 and 9, for the successive
sets of 100 images in the test set.  In total, there are 300 images in 
the test set.  Discrimination with the 7-dimensional data is far purer
than with the 72-dimensional data.}
\begin{center}
Original data, 72-dimensional \\
\begin{tabular}{r|rrr} \hline
(found) class  &  2   &   4  & 9 \\  \hline
(real) class 2 &  22  & 51  & 27 \\
             4 &   6  & 85  &  9 \\
             9 &   5  & 11  & 84 \\ \hline
\end{tabular}
 \ \  \\
\bigskip
 \ \  \\
7-dimensional factor space: factors 1--7 \\
\begin{tabular}{r|rrr} \hline
(found) class  &  2   &   4  & 9 \\  \hline
(real) class 2 &  60  & 19  & 21 \\
             4 &   1  & 97  &  2 \\
             9 &   2  &  7  & 91 \\ \hline  
\end{tabular}
\end{center}
\label{tabtabtab}
\end{table}

\begin{figure}[htbp] 
   \centering
   \includegraphics[width=14cm]{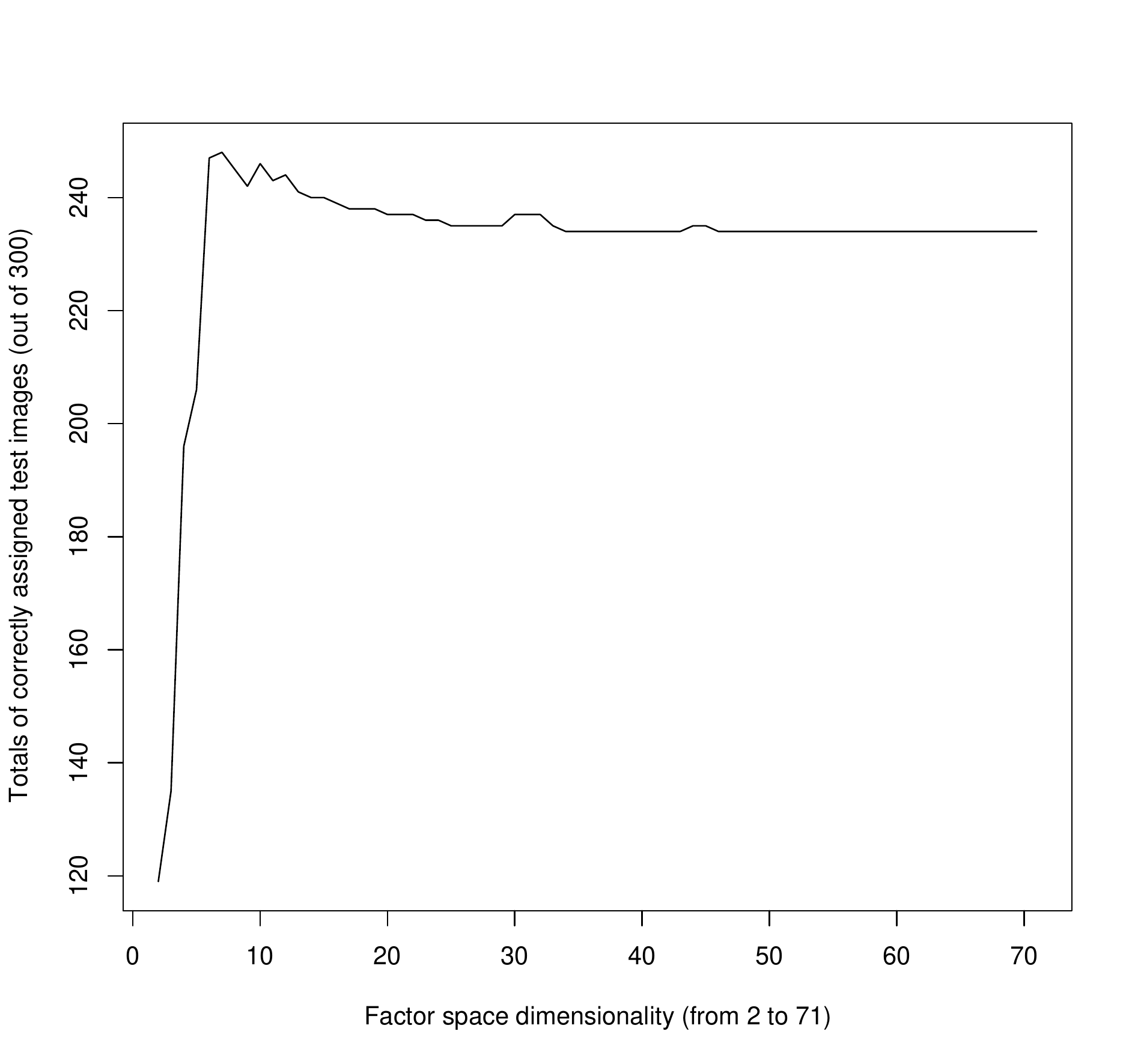}
\caption{Totaled correct assignments for best fit Euclidean (factor) subspaces,
in dimensions 2 to 71.}
\label{histCA}
\end{figure}

The assignments found are shown in Table \ref{tabtabtab}.  Justification 
for the choice of the 7-dimensional best Euclidean reduced-dimensionality 
embedding is derived from Figure \ref{histCA}.   For the original, full 
72-dimensionality data (Table \ref{tabtabtab}), the correct assignments were
respectively for the three classes 22, 85 and 84, all out of 100 images.  
For the best Euclidean embedding, viz.\ the 7-dimensional one, furnished 
by correspondence analysis, the correct assignments were
respectively for the three classes 60, 97 and 91, all out of 100 images.

As can be seen, the analysis of the low dimensional, correspondence analysis 
result is impressive relative to analysis of the input data.
In this 7-dimensional factor space, we ask next what are the predominant
features.  Looking at histograms of the contributions (i.e., sum of 
mass times projected distance squared from the origin) by features 
to factors 1 through 7, a threshold of 0.1 is either a natural one, or 
else is a reasonable choice.  This furnishes the following 
predominant features as follows:

\begin{itemize}

\item wavelet scale 5, 4th order moment

\item curvelet band 1, 2nd order moment

\item curvelet band 7, 3rd and 4th order moments

\item curvelet band 8, 4th order moment

\item curvelet band 11, 4th order moment, for two of the factors

\item curvelet band 12, 4th order moment
 
\item curvelet band 16, 4th order moment, for two of the factors

\item curvelet band 19, 2nd and 4th order moments, in the case of the 4th for 
two of the factors 

\end{itemize}

What is apparent here is that the 4th order moment has clear 
discriminatory power, although it is not unique in this capability.  
It is also apparent that the curvelet transform is very powerful 
in furnishing discriminatory features.  

\section{Conclusions}

Second order moment, or energy, has traditionally been used in image 
retrieval, including retrieval supported by multiple resolution transforms.
For example, see Fatemi-Ghomi (1997), who uses energy at multiple scales; 
Kubo et al.\ (2003), using energy and standard deviation at multiple scales; 
and Kokare et al.\ (2005), using up to second order autocorrelations, again
at multiple scales. In Starck et al.\ (1998b) it 
was shown how, under a Gaussian model assumption, the second order moment
could be viewed as a Shannon entropy.  For some types of imagery, the 
second order moment is a useful discriminator.  But not for all, and while
wavelet coefficients may be long-tailed they are not -- as we have 
shown in this article -- always so.  

We have shown that taking the second, third and fourth moments as features, 
at multiple resolution scales, may enhance discrimination between
images in the image set used.  Clearly the first order moment is of no 
use to us in the context of such transforms.  

Our results point very clearly towards the importance of 4th order 
moments of curvelet transform coefficients.  

These moments provide a 
proxy or substitute for the appropriate entropy to characterize the 
information, from among the mixture of appropriate entropies.  
Starck et al.\ (2004, 2005) take this work in the direction of 
characterizing non-Gaussian signatures (e.g., degree of clustering,
filamentarity, sheetedness and voidness) in the data.  For this, we  
use a battery of multiresolution transforms (discussing, inter alia, 
the product of kurtosis values yielded by different multiresolution 
transforms at given scales or bands).  

In this article we have 
applied this perspective to new classes of image and found 
excellent results in doing so.

\end{document}